\pdfoutput=1

\documentclass[11pt]{article}

\usepackage[preprint]{acl}

\usepackage{times}
\usepackage{latexsym}

\usepackage{booktabs}
\usepackage{rotating}
\usepackage{tabularx}

\usepackage{graphicx}    
\usepackage{subcaption}  

\renewcommand{\paragraph}[1]{\par\noindent\textbf{#1}}
\usepackage{xspace}
\newcommand{\F}{F$_1$\xspace}

\usepackage{amsfonts} 
\usepackage{amsmath}  

\usepackage{multirow}

\usepackage[T1]{fontenc}

\usepackage[utf8]{inputenc}

\usepackage{microtype}

\usepackage{inconsolata}

\title{Self-Adaptive Paraphrasing and Preference Learning for\\Improved Claim Verifiability}

\author{Amelie W\"uhrl$^{1,2}$  \and Roman Klinger$^2$ \\
  $^1$University of Stuttgart, Germany  
  $^2$University of Bamberg, Germany \\
  \texttt{firstname.lastname@uni-bamberg.de}\\
}

\begin{document}
\maketitle
\begin{abstract}
In fact-checking, structure and phrasing of claims critically influence a model's ability to predict verdicts accurately. 
Social media content in particular rarely serves as optimal input for verification systems, which necessitates pre-processing to extract the claim from noisy context before fact checking. Prior work suggests extracting a claim representation that humans find to be checkworthy and verifiable. This has two limitations: (1) the format may not be optimal for a fact-checking model, and (2), it requires annotated data to learn the extraction task from. 
We address both issues and propose a method to extract claims that is not reliant on labeled training data. Instead, our self-adaptive approach only requires a black-box fact checking model and a generative language model (LM). 
Given a tweet, we iteratively optimize the LM to generate a claim paraphrase that increases the performance of a fact checking model. By learning from preference pairs, we align the LM to the fact checker using direct preference optimization. 
We show that this novel setup extracts a claim paraphrase that is more verifiable than their original social media formulations, and is on par with competitive baselines. For refuted claims, our method consistently outperforms all baselines.
\end{abstract}


\section{Introduction}
In fact-checking, structure, length and the overall claim representation impact models' ability to reliably predict a verdict. Despite increased resources and modeling efforts dedicated to user-generated medical content and organically occurring medical claims on social media, a performance gap remains in fact-checking across different types of claims~\citep{kim-etal-2021-robust, wuhrl-klinger-2022-entity}. Possibly, this is because naturally occurring claims are more complex and longer, and contain multiple, inter-related facts compared to claims in other fact verification settings~\citep{sarrouti-etal-2021-evidence-based, Zuo2022}.
Since models do not transfer robustly to colloquial claims~\citep{kim-etal-2021-robust}, we hypothesize that this is because such claims are not optimal input for fact checking models.
Consider this tweet stating: `Just saw someone claiming that sipping on boiled garlic water is the magic cure for COVID-19. Anyone else heard this one?'. 
The checkworthy claim `Drinking boiled garlic water cures COVID-19' is embedded in context, potentially distracting the fact checking model and deteriorating its performance.
Given that the same model performs robustly on the concise, extracted version of the claim~\citep{wuhrl-klinger-2022-entity}, we presume that adapting these properties could enhance the fact-checking process for colloquial claims. Instead of modifying the model to accommodate colloquial input, we therefore propose to refine the input itself for better alignment with the model.

\begin{figure}[]
        \centering
        \includegraphics[width=0.35\textwidth]{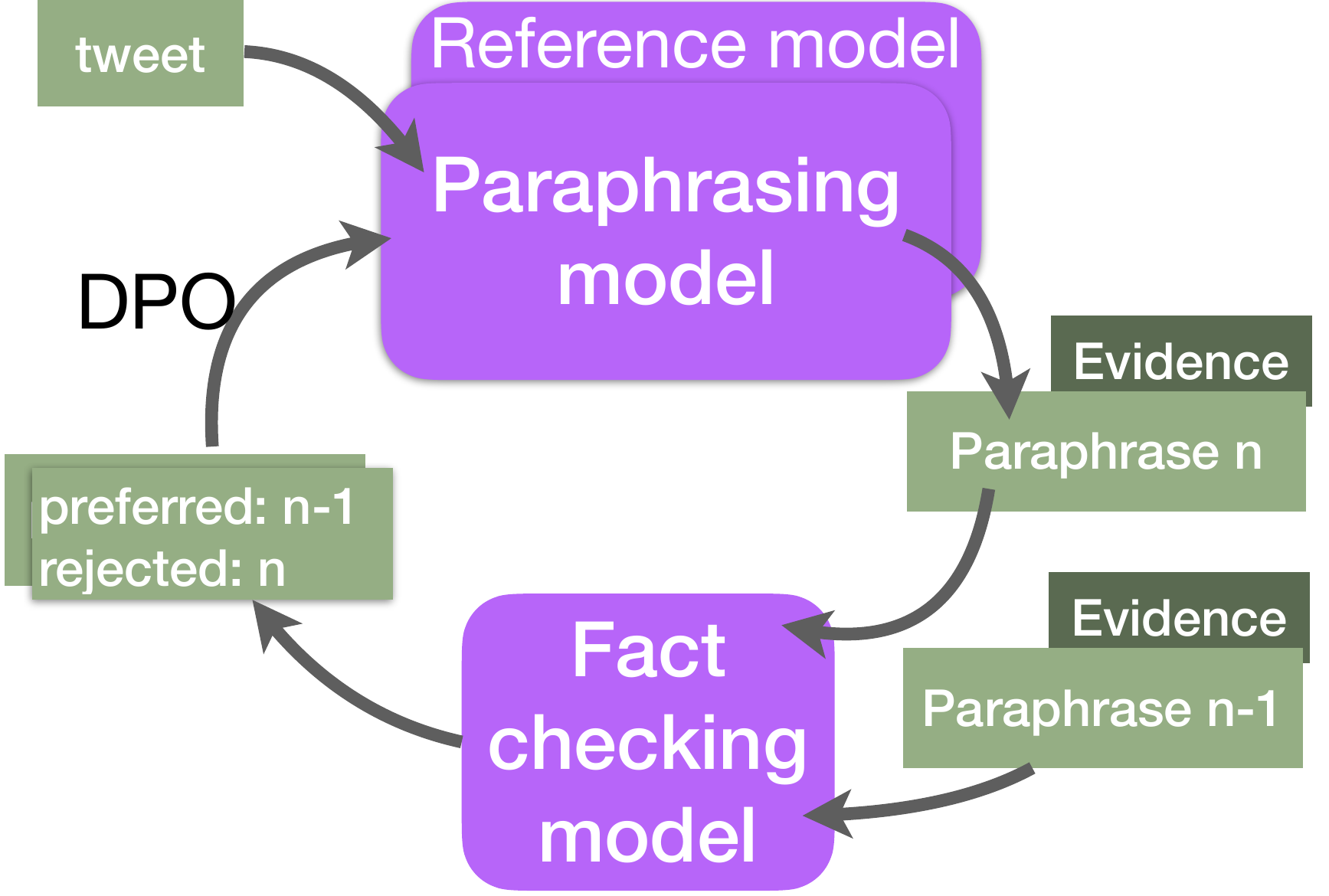}
        \caption{Illustration of the self-adaptive optimization cycle using direct preference optimization (DPO) guided by a fact-checking (FC) model.}
        \label{fig:dpo-illustration}
    \end{figure}

Claim extraction offers an intuitive solution.
Prior work explores extracting claims from long documents~\citep{deng-etal-2024-document} or noisy contexts~\citep{sundriyal-etal-2023-chaos}, for instance to identify checkworthy claims in discourse or to aligning the expert terminology of individual claim components with the language used in evidence documents~\citep{wuehrl-etal-2023-entity}.
Related tasks include claim detection, which identifies claim documents or sentences in argument mining~\citep[i.a.]{Lippi2015,gangi-reddy-etal-2022-newsclaims,zaberer-etal-2023-political}, and checkworthiness detection, which identifies claims that require fact-checking~\citep[i.a.]{hassan-et-al-2017_claimbuster,wright-augenstein-2020-claim,majer-snajder-2024-claim}.

Importantly, prior work extracts a claim representation that humans find to be checkworthy and verifiable. This has two limitations: (1) it requires annotated data to learn the extraction task from and (2) even with gold labeled data, the format may not be optimal for a fact-checking model. 
Optimizing the extracted claim for a downstream model, however, can not be learned end-to-end. 
Reinforcement learning (RL) is one way to address such problems as it allows learning from environment feedback and a reward objective as opposed to from labeled data.
In conjunction with the increased access to highly fluent generative language models, RL methods that align model outputs to preferences, have grown in popularity.
For instance, \citet{bing-et-al-2023_rl-based-counter-misinfromation-generation} and \citet{ziegenbein-etal-2024-llm} explore RL to generate counter misinformation responses and to increase argument appropriateness.
To address the unstable and computationally expensive nature of RL-based optimization, \citet{rafailov-et-al-2024_dpo} recently propose direct preference optimization (DPO). DPO is an alignment algorithm to optimize a generation policy, i.e., a language model. Learning from preference pairs, a large language model (LLM) is trained to to assign high probabilities to preferred prompt completions, while assigning lower probabilities to tokens of rejected completions. 

We build on this and propose a self-adaptive, iterative framework for claim extraction. We generate claim paraphrases that aligns the verification input based on a preference signal coming from a fact checking model.
Figure~\ref{fig:dpo-illustration} illustrates the DPO-based alignment cycle. 
Starting with a colloquial claim from social media and an off-the-shelf LLM, we iteratively extract the claim through paraphrasing. Each iteration updates the model using direct preference optimization, leveraging the feedback from the fact-checking model to steer the LLM generations towards a claim paraphrase that enhances verifiability.

We investigate two research questions: [RQ1] How effective is self-adaptive, DPO-based claim paraphrasing to enhance verifiability? and [RQ2] Which claim properties emerge throughout the self-adaptive paraphrasing process?
We show that this novel setup extracts a claim representation that is more verifiable than their original social media formulations, and is on par with competitive baselines. For refuted claims, our method consistently outperforms all baselines.
A key finding form our analysis is that self-adapted claims are very concise compared to their social media variants and even shorter than human-written claims.

\section{Methods}
\label{methods}

Given a fact checking model trained to predict a fact checking verdict for claim-evidence pairs, and a social media post that contains a medical claim, our goal is to extract a paraphrase of the claim that constitutes the best, i.e., most checkable, input for the fact checking model. 
Figure~\ref{fig:dpo-illustration} illustrates the iterative process we suggest to optimize the input claim.
Given a generative language model, we task the model to extract the claim from a social media document using zero-shot prompting. 
We pass the extracted claim along with its evidence document\footnote{As we focus on claim extraction, we presume an oracle setting and use the gold annotated evidence from a dataset.} to the fact checking model which predicts an entailment-style fact checking verdict, indicating if the evidence \textsc{supports} or  \textsc{refutes} the claim or if their relation is \textsc{neutral}.
For each claim, we compare the prediction of the fact checking model with the prediction for the claim-evidence pair from the previous iteration. This way, we obtain a preference pair: the claim which was more reliably checked, is the preferred claim; the other claim is the rejected one.
Using these preference pairs, we update the language model by fine-tuning it with the DPO loss. This aims to align the language model generations to the fact checking model's expected input, while constraining the update using a reference language model to avoid phenomena such as reward hacking.
After fine-tuning, we use the updated model to generate new claim paraphrases given the social media posts and continue the process for $n$ iterations.

\paragraph{Direct preference optimization.}
Direct preference optimization~\cite{rafailov-et-al-2024_dpo} is an algorithm to optimize a generation policy, i.e., a language model, by learning from preference pairs.
Intuitively speaking, given a preferred and a rejected completion to a prompt, the LLM is trained to assign high probabilities to the preferred output, while assigning rejected completions lower probabilities. 
In formal terms, given a preference dataset $D$ which consists of triplets $x, y_w, y_l$, where $x$ is a prompt with a chosen ($y_w$) and rejected ($y_l$) completion, we fine-tune a language model $\pi_\theta$ with the loss function $\mathcal{L}_\mathrm{DPO}$. The optimization is KL-constrained against a reference model $\pi_\mathrm{ref}$ and scaled by the parameter $\beta$:
\begin{multline}
\mathcal{L}_\mathrm{DPO}(\pi_\theta \; ; \; \pi_\mathrm{ref}) = \\
-\mathbb{E}_{(x, y_w, y_l) \sim D} \Big[ 
\log \sigma \Big( \beta \log \frac{\pi_\theta(y_w \mid x)}{\pi_\mathrm{ref}(y_w \mid x)} \\
- \beta \log \frac{\pi_\theta(y_l \mid x)}{\pi_\mathrm{ref}(y_l \mid x)} \Big) \Big].
\end{multline}

\paragraph{Preference pairs.} 
Given two claim paraphrases and their respective predictions from a fact checking model, we prefer the one with the correct label. If both predictions match the gold label, we choose the one with higher label confidence. If neither is correct but share the same incorrect label, we prefer the one with lower confidence. If both are incorrect but differ, we select randomly, unless one prediction is \textsc{neutral}, in which case it is preferred.

\paragraph{Data.}
\label{data}
We aim to understand how to optimize claim extraction from medical social media posts. Thus, we require social media texts that convey medical claims which we generate using a large language model. 
Given a seed claim $c_s$ from a dataset for biomedical fact checking $D$, we generate a tweet-style paraphrase  $c_\mathrm{tw}$. To obtain a diverse set of synthetic tweets, we prompt the model using randomly generated personas. We provide details on the personas in Appendix~\ref{appendix:data-personas}. Table~\ref{tab:claim-examples} shows an example seed claim, its evidence and synthetic tweet. 
We use synthetic tweets for two reasons. First, existing biomedical fact-checking datasets lack tweets paired with extracted claims which prevents comparison of DPO-paraphrased claims and human-written ones. Second, fact checking relies on claim-specific evidence. Social media data frequently contains multiple claims within one document~\cite{wuehrl-etal-2024-makes}. This tasks the model to identify relevant claims before learning to optimally phrase them.


\section{Experiments}
\label{experiments}
\subsection{Experimental Setting}
\label{ex-setting}
We run the optimization loop (Fig.~\ref{fig:dpo-illustration}) as described in Sec.~\ref{methods} with the components outlined in Sec.~\ref{ex-components} for a total of 10 iterations.
The fact checking performance serves as a proxy to evaluate the paraphrases. After each DPO update, the model generates new paraphrases for the test portion of the dataset. We pass the claim paraphrases together with their evidences to the fact checking model and evaluate its performance using precision, recall and F$_1$.

We gauge how self-adaptive extraction compares to alternative setups, namely no extraction, either leaving the claim embedded in a social media post or using the seed claim which is isolated by nature, and zero-shot extraction.  
Thus, we compare the fact checking performance for the following claim inputs: seed claim $c_s$ (the upper bound), synthetic tweet $c_\mathrm{tw}$ (baseline 1), zero-shot-extracted checkworthy claim (0-cw) and zero-shot-extracted core claim (0-ex) (baselines 2 and 3, respectively). Refer to Appendix~\ref{app:ex-details_baselines} for details on the baselines.

\subsection{Components}
\label{ex-components}
\paragraph{Dataset.}
We use the HealthVer dataset~\citep{sarrouti-etal-2021-evidence-based} for evidence-based fact-checking of health-related online claims. The dataset consists of 14,330 claim-evidence pairs. The claims serve as our seed claims $c_s$. Using \texttt{Llama-3-8B-Instruct}, we generate synthetic tweets $c_\mathrm{tw}$ that convey $c_s$ in the style of a social media post. Refer to Appendix~\ref{appendix:data} for prompting details.

\paragraph{Fact checking.}
We frame fact checking as an entailment or Natural Language Inference (NLI) task. Each instance is a premise-hypothesis pair. The claim is the hypothesis, while the evidence is the premise. The model predicts whether the claim is \textsc{entailed} or \textsc{contradicted} by the evidence or if there is a \textsc{neutral} relation between the two. 
As the fact checking model, we use \texttt{mDeBERTa}, a RoBERTa-based medium-sized model, trained for multilingual NLI\footnote{\url{https://huggingface.co/MoritzLaurer/mDeBERTa-v3-base-mnli-xnli}}. We choose to experiment with this model for two reasons: (a) it omits the need for task-specific training data and (b) it is lightweight and computationally efficient. Appendix~\ref{app:ex-details_fc} outlines the implementation details.

\paragraph{Paraphrasing.}
Our base and reference model for paraphrasing is \texttt{Llama-3-8B-Instruct}, which we update in each iteration. The model learns from preference pairs, i.e., a chosen and a rejected completion to the following prompt: \texttt{Your task is to extract the checkworthy claim from a piece of text. Here is the text: <$c_\mathrm{tw}$>.} We instruct the model to output json, providing the system prompt: \texttt{You are a fact checking assistant.}
For efficient fine-tuning, we use a LoRA adapter~\citep{hu2022lora} and train for two epochs using the DPO loss. Appendix~\ref{app:ex-details_dpo} provides the implementation details.
We use the train-dev-test split as provided in the HealthVer dataset.

\subsection{Results}
\label{results}
Our goal is to understand how effective self-adaptive, DPO-based claim paraphrasing is in enhancing verifiability (RQ1). Table~\ref{tab:fc-avg-results} shows the fact checking results (weighted \F) across claim inputs.
Across the iterations, the performance increases slightly (from .40\F to .43\F). This indicates that the self-adapted claim paraphrases present a more suitable input for the fact checking model as we keep updating the model.
Compared to the zero-shot baselines, the iterative processes outperforms 0-cw and achieves comparable performance to 0-ex. The upper bound achieves an \F-score of .46. 
The least suitable input for the fact checker is the unchanged tweet (.34\F), indicating that claim extraction is always beneficial.

\begin{figure}[]
        \centering
        \includegraphics[scale=.5]{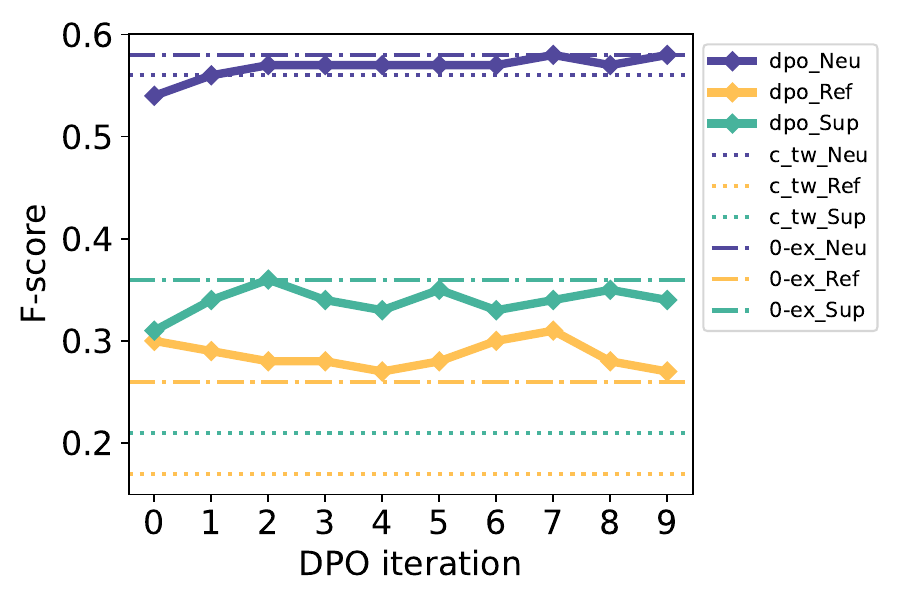} 
        \caption{Per class fact checking performance (F scores) across varying claim inputs.
        }
    
        \label{fig:results_per-class}
    \end{figure}

Figure~\ref{fig:results_per-class} plots the per class \F-scores for self-adaptive paraphrases compared to the strongest baseline (0-ex) and inputting an unextracted claim, i.e., the tweet. 
For \textsc{Neutral}, performance increases mostly consistently across iterations (max\F: .60, min\F: .54).
and \textsc{Supported} claims, performance increases until iteration 2 and fluctuates afterwards (max\F: .36, min\F: .31).
The performance for \textsc{Refuted} claims does not shows any consistency in the performance, with \F-scores fluctuating between .27 and .31.
Compared to the baseline, all inputs lead to comparable performances for \textsc{neutral} claims. For \textsc{supported} claims, the zero-shot extraction mostly outperforms the self-adaptive claims. For \textsc{refuted} claims, the self-adaptive claims outperform all baselines.

Table~\ref{tab:claim-examples} shows examples of the generated paraphrases. While the initial iterations show substantial changes, paraphrases stagnate after, reflecting the plateauing fact checking performance.
\begin{table}[]
\centering\small
\setlength{\tabcolsep}{1.5pt}
\begin{tabular}{cccc|cccccccccc}
\toprule
\multicolumn{4}{c}{}& \multicolumn{10}{c}{DPO iteration}\\
\cmidrule(l){5-14}
 $c_s$ & $c_\mathrm{tw}$ & 0-ex & 0-cw &0 &1&2&3&4&5&6&7& 8& 9 \\
\cmidrule(r){1-4} \cmidrule(l){5-14}
.46 &.34 &.43 &.40 &.40&.42&.42&.42&.41&.43&.42&.43&.42&.42\\
\bottomrule
\end{tabular}
\caption{Fact checking results (weighted \F) across claim inputs.}
\label{tab:fc-avg-results}
\end{table}

\section{Analysis}
To understand which claim properties emerge throughout the self-adaptive paraphrasing process and how the claims compare to the seed claims (RQ2), we perform two analyses. 

As we hypothesize that concise claims are more robustly verified, we first analyze claim length.
On average, tweets ($c_\mathrm{tw}$) consist of 41 words. For the first two iterations, claim length decreases dramatically, on average, down to 14.9 words. After that, claim length stagnates, indicating minimal changes in later paraphrases. Notably, the self-adapted claims are shorter than the seed claims. Table~\ref{tab:claim-len} shows the results in detail. 

Second, we compare similarity between each iteration's paraphrases and seed claims using BLEU, METEOR, and translation error rate (TER), which measures the number of edits required (see Table~\ref{tab:analysis-sim-metrics}). 
All metrics show increasing similarity over the first two iterations, before stagnating for the remaining rounds.
However, the absolute scores for all metrics indicate only modest similarity between self-adapted and seed claims.
This is not necessarily bad, instead, it supports our hypothesis that claims optimized for a fact checking model may differ from human-formulated claims.

\section{Conclusion}
We propose a self-adaptive framework for extracting online biomedical claims. To optimize fact verification inputs for fact checking, we iteratively fine-tune a LLM using preference learning. The preference signal comes from a fact checking model to generate a claim paraphrase that is more verifiable for the fact checker.
Our method increases the verifiability of claims compared to their original social media formulations. However, zero-shot extraction presents a competitive baseline
While zero-shot extraction is a competitive baseline for \textsc{Supported} and \textsc{Neutral} claims, our method consistently outperforms all baselines for \textsc{refuted} claims.

\section*{Limitations}
While instantiating the individual components is limited to one set of models and focused on one dataset, we choose the starting components in a way that they are general enough to gauge the capability of our method. Specifically, we work with a state-of-the-art large language model and a general-domain approach to fact checking using the NLI, instead of using a highly specialized model for biomedical fact checking. 
This being said, all components may of course be optimized, for example by adapting the fact checking model, to improve the overall performance. However, since we are interested in the effect of varying the input claims, the overall performance is somewhat negligible. It is more important to gauge the performance delta between  iterations and inputs.

Synthetic data may not be fully representative of the diverse nature of online discourse. While we prompt with different personas to increase variety, we observe in a manual inspection that the synthetic tweets frequently use similar paraphrases such as embedding the seed claim in tweets starting with ``Just learned that \ldots" or posing ``Did you know that\ldots" type questions to convey the claim. Presumably this could be a result of instruction tuning, leading the model to use such rhetoric instead of spreading unverified claims. 
In the future, we have to explore this method for other datasets and domains to understand its capabilities for highly diverse checkworthy content.

We constrain the updates to the paraphrasing model using a reference model, which is common practice for LLM alignment methods both in reinforcement learning for human feedback (RLHF) and direct preference optimization (DPO). This is intended to avoid reward hacking and keep outputs coherent. However, we hypothesize that this is one of the reasons the paraphrases stagnate after the initial iterations.
In the future, we aim to investigate the optimization process without this constraint as a way to understand which claim properties the fact checking model may exploit when unguided. Perhaps this advances our understanding of the weaknesses of the fact checking model, while also shedding light on which claim elements remain when removing a readability constraint.
On a similar note, exploring other ways to address the stagnating paraphrases --while out of scope for our prove-of-concept study-- is crucial. Considering how sensitive LLMs are with respect to prompt variation, future work has to investigate the effect of alternative extraction prompts, specifically to `encourage' the model to generate paraphrases that move away from the original wording in the social media post. Alternatively, we may adapt sampling strategies or other generation parameters to allow for more variation in the output.

\section*{Acknowledgements}
This work was conducted and funded as part of the \texttt{CEAT} project (DFG, KL~2869/1-2.).

\bibliography{anthology,custom}

\begin{thebibliography}{18}
\expandafter\ifx\csname natexlab\endcsname\relax\def\natexlab#1{#1}\fi

\bibitem[{Deng et~al.(2024)Deng, Schlichtkrull, and Vlachos}]{deng-etal-2024-document}
Zhenyun Deng, Michael Schlichtkrull, and Andreas Vlachos. 2024.
\newblock \href {https://doi.org/10.18653/v1/2024.acl-long.645} {Document-level claim extraction and decontextualisation for fact-checking}.
\newblock In \emph{Proceedings of the 62nd Annual Meeting of the Association for Computational Linguistics (Volume 1: Long Papers)}, pages 11943--11954, Bangkok, Thailand. Association for Computational Linguistics.

\bibitem[{Gangi~Reddy et~al.(2022)Gangi~Reddy, Chinthakindi, Wang, Fung, Conger, ELsayed, Palmer, Nakov, Hovy, Small, and Ji}]{gangi-reddy-etal-2022-newsclaims}
Revanth Gangi~Reddy, Sai~Chetan Chinthakindi, Zhenhailong Wang, Yi~Fung, Kathryn Conger, Ahmed ELsayed, Martha Palmer, Preslav Nakov, Eduard Hovy, Kevin Small, and Heng Ji. 2022.
\newblock \href {https://doi.org/10.18653/v1/2022.emnlp-main.403} {{N}ews{C}laims: A new benchmark for claim detection from news with attribute knowledge}.
\newblock In \emph{Proceedings of the 2022 Conference on Empirical Methods in Natural Language Processing}, pages 6002--6018, Abu Dhabi, United Arab Emirates. Association for Computational Linguistics.

\bibitem[{Hassan et~al.(2017)Hassan, Arslan, Li, and Tremayne}]{hassan-et-al-2017_claimbuster}
Naeemul Hassan, Fatma Arslan, Chengkai Li, and Mark Tremayne. 2017.
\newblock \href {https://doi.org/10.1145/3097983.3098131} {Toward automated fact-checking: Detecting check-worthy factual claims by claimbuster}.
\newblock In \emph{Proceedings of the 23rd ACM SIGKDD International Conference on Knowledge Discovery and Data Mining}, KDD '17, page 1803–1812, New York, NY, USA. Association for Computing Machinery.

\bibitem[{He et~al.(2023)He, Ahamad, and Kumar}]{bing-et-al-2023_rl-based-counter-misinfromation-generation}
Bing He, Mustaque Ahamad, and Srijan Kumar. 2023.
\newblock \href {https://doi.org/10.1145/3543507.3583388} {Reinforcement learning-based counter-misinformation response generation: A case study of covid-19 vaccine misinformation}.
\newblock In \emph{Proceedings of the ACM Web Conference 2023}, WWW '23, page 2698–2709, New York, NY, USA. Association for Computing Machinery.

\bibitem[{Hu et~al.(2022)Hu, Shen, Wallis, Allen-Zhu, Li, Wang, Wang, and Chen}]{hu2022lora}
Edward~J Hu, Yelong Shen, Phillip Wallis, Zeyuan Allen-Zhu, Yuanzhi Li, Shean Wang, Lu~Wang, and Weizhu Chen. 2022.
\newblock \href {https://openreview.net/forum?id=nZeVKeeFYf9} {Lo{RA}: Low-rank adaptation of large language models}.
\newblock In \emph{International Conference on Learning Representations}.

\bibitem[{Kim et~al.(2021)Kim, Kim, Hong, and Kim}]{kim-etal-2021-robust}
Byeongchang Kim, Hyunwoo Kim, Seokhee Hong, and Gunhee Kim. 2021.
\newblock \href {https://doi.org/10.18653/v1/2021.naacl-main.121} {How robust are fact checking systems on colloquial claims?}
\newblock In \emph{Proceedings of the 2021 Conference of the North American Chapter of the Association for Computational Linguistics: Human Language Technologies}, pages 1535--1548, Online. Association for Computational Linguistics.

\bibitem[{Lippi and Torroni(2015)}]{Lippi2015}
Marco Lippi and Paolo Torroni. 2015.
\newblock \href {https://www.ijcai.org/Proceedings/15/Papers/033.pdf} {Context-independent claim detection for argument mining}.
\newblock In \emph{Proceedings of the 24th International Conference on Artificial Intelligence}, IJCAI'15, page 185–191. AAAI Press.

\bibitem[{Majer and {\v{S}}najder(2024)}]{majer-snajder-2024-claim}
Laura Majer and Jan {\v{S}}najder. 2024.
\newblock \href {https://doi.org/10.18653/v1/2024.fever-1.27} {Claim check-worthiness detection: How well do {LLM}s grasp annotation guidelines?}
\newblock In \emph{Proceedings of the Seventh Fact Extraction and VERification Workshop (FEVER)}, pages 245--263, Miami, Florida, USA. Association for Computational Linguistics.

\bibitem[{Rafailov et~al.(2024)Rafailov, Sharma, Mitchell, Ermon, Manning, and Finn}]{rafailov-et-al-2024_dpo}
Rafael Rafailov, Archit Sharma, Eric Mitchell, Stefano Ermon, Christopher~D. Manning, and Chelsea Finn. 2024.
\newblock Direct preference optimization: your language model is secretly a reward model.
\newblock In \emph{Proceedings of the 37th International Conference on Neural Information Processing Systems}, NIPS '23, Red Hook, NY, USA. Curran Associates Inc.

\bibitem[{Sarrouti et~al.(2021)Sarrouti, Ben~Abacha, Mrabet, and Demner-Fushman}]{sarrouti-etal-2021-evidence-based}
Mourad Sarrouti, Asma Ben~Abacha, Yassine Mrabet, and Dina Demner-Fushman. 2021.
\newblock \href {https://doi.org/10.18653/v1/2021.findings-emnlp.297} {Evidence-based fact-checking of health-related claims}.
\newblock In \emph{Findings of the Association for Computational Linguistics: EMNLP 2021}, pages 3499--3512, Punta Cana, Dominican Republic. Association for Computational Linguistics.

\bibitem[{Sundriyal et~al.(2023)Sundriyal, Chakraborty, and Nakov}]{sundriyal-etal-2023-chaos}
Megha Sundriyal, Tanmoy Chakraborty, and Preslav Nakov. 2023.
\newblock \href {https://doi.org/10.18653/v1/2023.findings-emnlp.439} {From chaos to clarity: Claim normalization to empower fact-checking}.
\newblock In \emph{Findings of the Association for Computational Linguistics: EMNLP 2023}, pages 6594--6609, Singapore. Association for Computational Linguistics.

\bibitem[{Wright and Augenstein(2020)}]{wright-augenstein-2020-claim}
Dustin Wright and Isabelle Augenstein. 2020.
\newblock \href {https://doi.org/10.18653/v1/2020.findings-emnlp.43} {Claim check-worthiness detection as positive unlabelled learning}.
\newblock In \emph{Findings of the Association for Computational Linguistics: EMNLP 2020}, pages 476--488, Online. Association for Computational Linguistics.

\bibitem[{Wuehrl et~al.(2023)Wuehrl, Grimminger, and Klinger}]{wuehrl-etal-2023-entity}
Amelie Wuehrl, Lara Grimminger, and Roman Klinger. 2023.
\newblock \href {https://doi.org/10.18653/v1/2023.fever-1.3} {An entity-based claim extraction pipeline for real-world biomedical fact-checking}.
\newblock In \emph{Proceedings of the Sixth Fact Extraction and VERification Workshop (FEVER)}, pages 29--37, Dubrovnik, Croatia. Association for Computational Linguistics.

\bibitem[{Wuehrl et~al.(2024)Wuehrl, Menchaca~Resendiz, Grimminger, and Klinger}]{wuehrl-etal-2024-makes}
Amelie Wuehrl, Yarik Menchaca~Resendiz, Lara Grimminger, and Roman Klinger. 2024.
\newblock \href {https://aclanthology.org/2024.eacl-long.124} {What makes medical claims (un)verifiable? analyzing entity and relation properties for fact verification}.
\newblock In \emph{Proceedings of the 18th Conference of the European Chapter of the Association for Computational Linguistics (Volume 1: Long Papers)}, pages 2046--2058, St. Julian{'}s, Malta. Association for Computational Linguistics.

\bibitem[{W{\"u}hrl and Klinger(2022)}]{wuhrl-klinger-2022-entity}
Amelie W{\"u}hrl and Roman Klinger. 2022.
\newblock \href {https://aclanthology.org/2022.argmining-1.18} {Entity-based claim representation improves fact-checking of medical content in tweets}.
\newblock In \emph{Proceedings of the 9th Workshop on Argument Mining}, pages 187--198, Online and in Gyeongju, Republic of Korea. International Conference on Computational Linguistics.

\bibitem[{Zaberer et~al.(2023)Zaberer, Pado, and Lapesa}]{zaberer-etal-2023-political}
Urs Zaberer, Sebastian Pado, and Gabriella Lapesa. 2023.
\newblock \href {https://aclanthology.org/2023.konvens-main.22} {Political claim identification and categorization in a multilingual setting: First experiments}.
\newblock In \emph{Proceedings of the 19th Conference on Natural Language Processing (KONVENS 2023)}, pages 219--228, Ingolstadt, Germany. Association for Computational Lingustics.

\bibitem[{Ziegenbein et~al.(2024)Ziegenbein, Skitalinskaya, Bayat~Makou, and Wachsmuth}]{ziegenbein-etal-2024-llm}
Timon Ziegenbein, Gabriella Skitalinskaya, Alireza Bayat~Makou, and Henning Wachsmuth. 2024.
\newblock \href {https://doi.org/10.18653/v1/2024.acl-long.244} {{LLM}-based rewriting of inappropriate argumentation using reinforcement learning from machine feedback}.
\newblock In \emph{Proceedings of the 62nd Annual Meeting of the Association for Computational Linguistics (Volume 1: Long Papers)}, pages 4455--4476, Bangkok, Thailand. Association for Computational Linguistics.

\bibitem[{Zuo et~al.(2022)Zuo, Mathur, Kela, Salek~Faramarzi, and Banerjee}]{Zuo2022}
Chaoyuan Zuo, Kritik Mathur, Dhruv Kela, Noushin Salek~Faramarzi, and Ritwik Banerjee. 2022.
\newblock \href {https://doi.org/10.1007/s41060-022-00310-7} {Beyond belief: A cross-genre study on perception and validation of health information online}.
\newblock \emph{International Journal of Data Science and Analytics}, 13(4):299--314.

\end{thebibliography}

\appendix
\section{Appendix}
\subsection{Data}
\label{appendix:data}
\paragraph{Synthetic tweets.}
For each seed claim $c_s$ in the HealthVer dataset~\citep{sarrouti-etal-2021-evidence-based}, we generate a tweet-style paraphrase $c_\mathrm{tw}$ that conveys the claim. Using \textsc{Llama-3-8B-Instruct}, we prompt the model as follows: \texttt{<persona system prompt> Your task is to write a Twitter post in which you paraphrase a claim or statement that I give you. Please paraphrase the statement so that it reads like one of your social media posts. Please format your reply as valid json: \{""post"": ""YOUR REPLY""\} Only output the json. Here is the statement: \textbf{< $c_s$ >}}
\paragraph{Persona system prompt.}
\label{appendix:data-personas}
To obtain a diverse set of synthetic tweets, we prompt the model using randomly generated personas which we include as a system prompt (see above). The persona system prompts are also LLM-generated (\textsc{Llama-3-8B-Instruct}). To this end, we randomly pick two demographic attributes and a profession to construct a persona: \texttt{You are <demographical attribute 1>. You are <demographical attribute 2>. You are <profession.>} We select the demographic attributes from the following list: [a teenager, a young adult, an adult, a senior citizen, a male social media user, a female  social media user, a non-binary social media user, American, Canadian, British, Indian, Chinese, Brazilian, Nigerian, Mexican, Japanese, Australian, British, French, German, Italian]. We select the profession from the following list: [ a retail cashier, a teacher, a receptionist, a customer service representative, a construction worker, a security guard, a barista, a truck driver, an electrician, a plumber, a carpenter, a mechanic, a HVAC technician, a welder, a software engineer, a nurse, an accountant, a marketing manager, a human resources manager, a graphic designer, a real estate agent, a pharmacist, a data scientist, a robotics engineer, a cybersecurity analyst, a marine biologist, a cryptographer, a neurosurgeon, an ethical hacker, a sommelier, an artisan cheesemaker, an astronaut, a high school student, a college student]. Subsequently, we provide the persona (example: You are a teenager. You are a non-binary social media user. You are a student.) as a system prompt and `ask' the model: \texttt{Who are you?} Here is an example output: ``Hey! I'm Zephyr, nice to meet you! I'm a non-binary teenager, which means I don't identify as strictly male or female. I'm still figuring out who I am and what I want to be, but I'm happy to be me, you know?"

\subsection{Experimental Details}
\label{app:ex-details}

\subsubsection{Fact checking}
\label{app:ex-details_fc}
We use the \texttt{transformers} library and provide the model with tokenized premise-hypothesis pairs. We convert the model output into probabilities for each class (\textsc{entailment}, \textsc{neutral}, \textsc{contradiction}) represented by the logits using Softmax. 
We run the experiments on a single Nvidia GeForce RTX A6000 GPU. Inference for all instances in the HealthVer data takes approx. 10 minutes.

\subsubsection{Model Training with DPO}
\label{app:ex-details_dpo}
We use the \texttt{transformers} and the \texttt{trl} library which implements the DPO loss function in the DPOTrainer. We train with the following parameters: num\_train\_epochs=2, per\_device\_train\_batch\_size=12, per\_device\_eval\_batch\_size=4,   gradient\_accumulation\_steps=1, gradient\_checkpointing=True, optim="adamw\_torch\_fused",   learning\_rate=5e-5, max\_grad\_norm=0.3, warmup\_ratio=0.1, lr\_scheduler\_type="cosine", logging\_steps=25, save\_steps=500, save\_total\_limit=2, eval\_strategy="steps", eval\_steps=700, bf16=True, beta=0.1, loss\_type="sigmoid"

To fine-tune the paraphrasing model in each iteration, we use a LoRA adapter which we train using the DPO loss. We configure the LoRA adapter as follows: lora\_alpha=128, lora\_dropout=0.05, r=256, bias="none", target\_modules="all-linear", task\_type="CAUSAL\_LM".

One DPO update (training for 2 epochs) takes approx. 2~hours and 45~minutes.

\subsubsection{Zero-shot Baselines}
\label{app:ex-details_baselines}
For the zero-shot extraction baselines, we use two prompt variants. One of them specifies to extract the \emph{core} claim, whereas the other specifies to extract the \emph{checkworthy} claim from the tweet. We refer to them as 0-ex and 0-cw, respectively.

The 0-ex prompt consists of the system prompt \texttt{You are a helpful, highly skilled assistant.} and the task prompt \texttt{Your task is to extract the core claim from a piece of text. Please format your reply as valid json: \{""post"": ""YOUR REPLY""\} Only output the json. Here is the text: <  $c_\mathrm{tw} $>}.

The 0-cw prompt consists of the system prompt \texttt{You are an experienced fact checker.} and the task prompt \texttt{Your task is to extract the checkworthy claim from a piece of text. Please format your reply as valid json: \{""post"": ""YOUR REPLY""\} Only output the json. Here is the text: <  $c_\mathrm{tw} $>}.

\subsection{Analysis}
\label{app:analysis}

Table~\ref{tab:claim-examples} showcases three seed claims along with their evidence pieces, synthetic tweets and paraphrases across the self-adaptive claim optimization process.

Table~\ref{tab:claim-len} shows the average claim lengths (in words) for the seed claims, synthetic tweets and paraphrases across the self-adaptive claim optimization process.

Table~\ref{tab:analysis-sim-metrics} shows the average BLEU and METEOR score and average translation error rate (TER) for the paraphrases we obtain after each iteration of DPO updates.

\begin{table*}[ht]
\centering\small

\begin{tabularx}{\textwidth}{lXXX}
\toprule
id & ex1 & ex2 & ex3 \\
\cmidrule(r){1-1} \cmidrule(rl){2-2} \cmidrule(rl){3-3} \cmidrule(l){4-4}
$c_s$ & Drinking boiled garlic water will cure COVID-19. 
& Social distancing is a voluntary practice to help stop the spread of COVID-19 
& there are few novel sars-cov-2 cases in malaria countries because of the use of the antimalarial drug hydroxychloroquine.\\

evidence & In conclusion, Allium sativum may be an acceptable preventive measure against COVID-19 infection to boost immune system cells and to repress the production and secretion of proinflammatory cytokines as well as an adipose tissue derived hormone leptin having the proinflammatory nature. 
& 
the centralized social distancing is necessary to minimize the pandemic spreading.
& HCQ should be prescribed as a part of treatment for critically ill COVID-19 patients, with possible outcome of saving lives.\\
\midrule
$c_\mathrm{tw}$ & Just saw someone claiming that sipping on boiled garlic water is the magic cure for COVID-19 Anyone else heard this one? Not buying it, folks! As a student, I've got my facts straight and there's no scientific evidence to back this up. Stay safe, stay informed, and let's not spread misinformation! \#COVID19 \#StaySafe \#FactCheck & 
Let's all do our part to keep each other safe! By choosing to stay a little farther apart, we can really help slow down the spread of COVID-19. It's all about looking out for one another, you know? \#SocialDistancing \#COVID19 \#StaySafe
& Did you know that the use of hydroxychloroquine in malaria-prone countries has led to a significant decline in new \#SARSCoV2 cases? It's amazing to see how this antimalarial drug has been repurposed to combat COVID-19! \#Hydroxychloroquine \#Malaria \#COVID19
\\
\cmidrule{2-4}
$\mathrm{dpo}_0$ & There is no evidence to support the claim that sipping on boiled garlic water is the magic cure for COVID-19. It is likely a form of misinformation that can be harmful if believed and spread. &
By choosing to stay a little farther apart, we can really help slow down the spread of COVID-19. It's all about looking out for one another, you know? \#SocialDistancing \#COVID19 \#StaySafe
& The claim that the use of hydroxychloroquine in malaria-prone countries has led to a significant decline in new SARS-CoV-2 cases is checkworthy.\\

$\mathrm{dpo}_1$ & Sipping on boiled garlic water is a magic cure for COVID-19 & Choosing to stay a little farther apart can help slow down the spread of COVID-19. 
& The use of hydroxychloroquine in malaria-prone countries has led to a significant decline in new \#SARSCoV2 cases
\\

$\mathrm{dpo}_2$ & Boiled garlic water is a magic cure for COVID-19 & By choosing to stay a little farther apart, we can really help slow down the spread of COVID-19.
& The use of hydroxychloroquine in malaria-prone countries has led to a significant decline in new \#SARSCoV2 cases.
\\

$\mathrm{dpo}_3$ & Sipping on boiled garlic water is the magic cure for COVID-19. & Staying a little farther apart can help slow down the spread of COVID-19. 
& \multicolumn{1}{c}{\raisebox{-1.5ex}{``}}
\\

$\mathrm{dpo}_4$ & \multicolumn{1}{c}{\raisebox{-1.5ex}{``}} & Choosing to stay a little farther apart can help slow down the spread of COVID-19. 
& \multicolumn{1}{c}{\raisebox{-1.5ex}{``}}
\\

$\mathrm{dpo}_5$ &\multicolumn{1}{c}{\raisebox{-1.5ex}{``}} & Staying a little farther apart can help slow down the spread of COVID-19.
& \multicolumn{1}{c}{\raisebox{-1.5ex}{``}}
\\

$\mathrm{dpo}_6$ & \multicolumn{1}{c}{\raisebox{-1.5ex}{``}}& By choosing to stay a little farther apart, we can really help slow down the spread of COVID-19.
& \multicolumn{1}{c}{\raisebox{-1.5ex}{``}}
\\

$\mathrm{dpo}_7$ &\multicolumn{1}{c}{\raisebox{-1.5ex}{``}} & \multicolumn{1}{c}{\raisebox{-1.5ex}{``}}
& \multicolumn{1}{c}{\raisebox{-1.5ex}{``}}
\\

$\mathrm{dpo}_8$ &\multicolumn{1}{c}{\raisebox{-1.5ex}{``}} & \multicolumn{1}{c}{\raisebox{-1.5ex}{``}}
& \multicolumn{1}{c}{\raisebox{-1.5ex}{``}}
\\

$\mathrm{dpo}_9$ & \multicolumn{1}{c}{\raisebox{-1.5ex}{``}} & Staying a little farther apart can really help slow down the spread of COVID-19.
& \multicolumn{1}{c}{\raisebox{-1.5ex}{``}}
\\

\bottomrule
\end{tabularx}
\caption{Example instance with paraphrases. The ditto marker \raisebox{-0.5ex}{``} indicates that the paraphrase is identical to the previous iteration.}
\label{tab:claim-examples}
\end{table*}

\begin{table*}[]
    \centering
    \setlength{\tabcolsep}{4pt}
    \begin{tabular}{lrrrrrrrrrrrr}
\toprule
 & $c_s$ &  $c_\mathrm{tw}$ &     0 &     1 &     2 &     3 &     4 &     5 &     6 &     7 &     8 &     9 \\
 \cmidrule(lr){2-2} \cmidrule(lr){3-3} \cmidrule(l){4-13}
   avg. \# words &    17.5 &  41.0 & 28.1 & 14.9 & 14.9 & 14.9 & 14.9 & 14.9 & 14.9 & 14.8 & 14.9 & 14.9 \\
\cmidrule(lr){2-2} \cmidrule(lr){3-3} \cmidrule(l){4-13}
       std & 10.2 &   9.4 & 13.0 &  5.8 &  5.8 &  5.8 &  5.8 &  5.8 &  5.8 &  5.8 &  5.7 &  5.8 \\
\bottomrule
\end{tabular}
    \caption{Mean claim length in words for seed claims $c_s$, tweets $c_\mathrm{tw}$ and across DPO iterations.}
    \label{tab:claim-len}
\end{table*}

\begin{table*}[]
\centering\small
\begin{tabular}{lcccccccccc}
\toprule
 & P\_$0$ & P\_$1$ & P\_$2$ & P\_$3$ & P\_$4$ & P\_$5$ & P\_$6$ & P\_$7$ & P\_$8$ & P\_$9$ \\
\cmidrule{2-11}
BLEU & $0.068$ & $0.094$ & $0.094$ & $0.094$ & $0.093$ & $0.093$ & $0.093$ & $0.093$ & $0.093$ & $0.094$ \\ 
\cmidrule{1-11}
METEOR & $0.314$ & $0.325$ & $0.324$ & $0.324$ & $0.324$ & $0.325$ & $0.324$ & $0.324$ & $0.323$ & $0.325$ \\ 
\cmidrule{1-11}
TER & $183.284$ & $96.230$ & $96.624$ & $96.485$ & $96.735$ & $96.846$ & $96.835$ & $96.601$ & $96.694$ & $96.557$ \\
\bottomrule
\end{tabular}
\caption{Average BLEU and METEOR score and average translation error rate (TER) for DPO paraphrases. P stands for paraphrase.}
\label{tab:analysis-sim-metrics}
\end{table*}

\end{document}